\documentclass{llncs}
\usepackage{subfigure,amsmath}
\usepackage{epsfig}
\usepackage{url,enumerate}
\usepackage{graphics,float}
\usepackage{multirow}
\usepackage{graphicx,color,url}
\usepackage[ruled,vlined]{algorithm2e}
\usepackage[shortlabels]{enumitem}
\usepackage{caption} 
\usepackage[export]{adjustbox}
\usepackage[perpage]{footmisc}
\usepackage{times}
\def\mydoubleq#1{``#1''}

\newcount\Comments  
\definecolor{darkgreen}{rgb}{0,0.6,0}
\definecolor{orange}{rgb}{1,0.4,0}
\definecolor{purple}{rgb}{1,0,1}
\newcommand{\kibitz}[2]{\ifnum\Comments=1{\textcolor{#1}{#2}}\fi}

\title{Combining Difficulty Ranking  with  Multi-Armed Bandits to Sequence Educational Content}
\author{Avi Segal\inst{1}, Yossi Ben David\inst{2},  Joseph Jay Williams\inst{3}, Kobi  Gal\inst{1}, Yaar Shalom\inst{1}}
\institute{Ben-Gurion University of the Negev, Israel \and Microsoft, Israel \and National University of Singapore, Singapore}
\begin{document}
\maketitle
\begin{abstract} 
As e-learning systems become more prevalent, there is a growing need for them to accommodate individual differences between students.
This paper addresses the problem of how to personalize educational content to students in order to maximize their learning gains over time.  
We present a new computational approach to this problem called MAPLE (Multi-Armed Bandits based Personalization for Learning Environments) that combines difficulty ranking with multi-armed bandits.
Given a set of target questions MAPLE estimates the expected learning gains for each question and uses an exploration-exploitation strategy to choose the next question to pose to the student. It maintains a personalized ranking over the difficulties of question in the target set which is used in two ways: First, to obtain initial estimates over the learning gains for the set of questions. Second, to update the estimates over time based on the student’s responses. We show in simulations that MAPLE was able to improve students' learning gains compared to approaches that sequence questions in increasing level of difficulty, or rely on content experts. When implemented in a live e-learning system in the wild, MAPLE 
showed promising results. This work demonstrates the efficacy of using stochastic approaches to the sequencing problem when augmented with information about question difficulty.
\end{abstract}
\section{Introduction}
As e-learning systems become more prevalent they are  accessed by students of varied backgrounds, learning styles and needs. There is thus a growing need for them to accommodate individual difference between students and adapt to their changing pedagogical needs over time. 
There are mainly two families of approaches for adapting educational content: offline approaches build models from data (e.g. ~\cite{segal2014edurank}) while online approaches balance exploration and exploitation (e.g ~\cite{clement2013multi}).

We  provide a novel algorithm for sequencing content in e-learning systems that combines both of these approaches. It integrates offline learning from students' past interactions with online mechanisms for sequencing questions to students in order to  maximize their  learning gains. 
This approach is based on the concept of \textit{zone of proximal development} \cite{vygotsky1987zone} where students are presented with challenges that are neither too easy nor too difficult, but are slightly beyond their current abilities.

Our algorithm, called MAPLE (Multi-Armed Bandits based Personalization for Learning Environments), 
 extends prior multi-armed bandits approaches in education by explicitly considering question difficulty when initializing the online behavior of the algorithm and when updating its behavior over time. MAPLE treats successful and failed question attempts differently and becomes \mydoubleq{conservative} (i.e. decreases exploration and increases weights of easier questions) when students fail under the (pedagogically guided) assumption  that repeated errors are detrimental to students' learning~\cite{pekrun2016academic}.

We first evaluated MAPLE in a simulation environment comparing its performance to a variety of sequencing algorithms, including an approach that sequenced questions according to educational expert guidelines, an approach based only on personalized difficulty ranking, and a multi-armed bandit approach without personalized difficulty ranking initialization. MAPLE outperformed all other approaches for average and strong simulated students while showing the need for further tuning for weak simulated students.

We then implemented MAPLE in the wild in an existing e-learning system in a school with 7th grade students. MAPLE's performance was compared to two other sequencing algorithms already implemented in the e-learning system: an approach that sequenced questions according to educational expert guidelines and a state of the art Bayesian Knowledge Tracing based algorithm \cite{david2016sequencing}. We found that our proposed approach showed  promising results compared to the existing educational expert approach and the BKT based approach. Moreover, students reported being more satisfied with the questions posed to them by MAPLE.

					
The contribution of this paper is two-fold. First, we present a novel algorithm for sequencing questions to students in e-learning that extends multi-armed bandits with personalized difficulty ranking information. Second, we show the potential of the algorithm   in simulations and in field trials in the wild. 
\section{Related Work}
Our work relates to past research on using historical data to sequence content to students, and to work on multi-armed bandits for online adaptation of educational content.

Several approaches within the educational artificial intelligence community have used computational methods for sequencing content to students.
Pardos and Heffernan~\cite{pardos2009determining} inferred order over question presented to students by predicting their skill levels using Bayesian Knowledge Tracing (BKT)~\cite{corbett1994knowledge}. They showed the efficacy of their approach on simulated data as well as on a test-set comprising random sequences of three questions. Ben David et al.~\cite{david2016sequencing} developed a BKT based sequencing algorithm. Their algorithm (which we refer to in this paper as YBKT) uses knowledge tracing to model students' skill acquisition over time and sequence questions to students based on their mastery level and predicted performance. It was shown to enhance student learning beyond sequencing designed by pedagogical experts.

Champaign et al.~\cite{champaign2010model} used a peer-based approach for content sequencing in an intelligent tutor system by computing similarities between students and choosing questions that provide best benefits for similar students, measured by similar average performance on past questions. Segal et al.~\cite{segal2014edurank} developed EduRank, a sequencing algorithm that combines collaborative filtering with social choice theory to produce personalized learning sequences for students. The algorithm constructs a “difficulty” ranking over questions for a target student by aggregating the ranking of similar students when sequencing educational content.

Multi-armed bandits provides a fundamental model for tackling the \mydoubleq{exploration-exploitation} trade off ~\cite{thompson1933likelihood,bubeck2012regret}. Xu et al. \cite{xu2016personalized} used bandits to identify which sequences of courses lead students to obtain maximal GPAs. Lan and Baraniuk~\cite{lan2016contextual} used sparse factor analysis with bandits to identify sequences of educational content that could maximize students performance on subsequent assessments. Lomas et al. \cite{lomas2016interface} showed how bandits can be used to search a large space of design decisions in creating educational games.  Williams et al.~\cite{williams2016axis} used Thompson Sampling to identify highly rated explanations for how to solve Math problems, and chose priors that assumed that every explanation was equally rated.
Clement et al.~\cite{clement2013multi} used human experts' knowledge to initialize  a multi-armed bandit algorithm called EXP4, that discovered which activities were at the right level to push students learning forward. In our work we do not rely on human experts, but rather use personalized difficulty rankings to guide the initial exploitation and update steps of our algorithm.

\section{Problem Formulation and Approach}
We consider an e-learning setting with a group of students $S$ and a set of practice questions $Q$. At each time step in the practice session, a computer agent $A$ needs to choose a question in $Q$ to present to the student. The agent sequencing problem  requires choosing  at each time step the next question to present to the student so as to maximize her learning gains over  the length of the practice session. The goal is to present students with challenging problems while ensuring a high likelihood that they will be able to solve these problems. 

Our approach to solving the problem, called MAPLE, maintains a  belief  distribution over  the expected learning gains to the student for solving each of the questions in $Q$.  This belief distribution is initialized with a personalized difficulty ranking over the questions in $Q$. The algorithm samples the next question to the student from this  distribution and updates it at each time step  given the student's performance on the question and its inferred difficulty  to the student.

MAPLE implements an exploration policy 
similar to the one that is used by the EXP4  algorithm~\cite{auer2002nonstochastic,clement2013multi}. This approach 
maintains a   belief distribution over questions that is  proportional to how much learning gain each question is expected to provide. Weights are decreased or increased  based on  how difficult a question is and whether the student successfully solves the question or not. When a student successfully solves a question, weights are adjusted to make harder questions more likely to be presented, and explore a broader range of questions. When a student fails to solve a question, weights are adjusted to make easier questions more likely to be presented, and explore a narrower set of questions.  The weights are initialized to reflect the inferred difficulty level for the student.


\begin{algorithm}[H]
 \KwData {Set of students $S$.
 \\Set of $N$ questions $Q$.
 \\Passing grade threshold $\eta$, exploration rate $\gamma$. 
 \\Normalization factors $\alpha_1,\ldots,\alpha_4$.
 \\For each student in $S$, a partial difficulty ranking $\succ_k$ over $T_k \subseteq Q$.
 \\Target student $s_i \in S$.}
 \KwResult {Next question $q_s \in Q$ to present to student $s_i$.}
{\bf{Step 1: Initialization:}} 
\begin{enumerate}[topsep=0pt]
\item For each student in $S$:\\
- Compute difficulty ranking $\succ_l$ for all $q_l \in Q$\\
- $w_1 > \ldots > w_N=initialize\_weights(\succ_l)$\\
\end{enumerate}
{\bf{Step 2: Get Next Question:}} 
\begin{enumerate}[topsep=0pt]
\setcounter{enumi}{1}
\item For each $j=1, \ldots , N$:\\
- Calculate new weight: $\pi_j = w_j(1-\gamma) + \epsilon_j\gamma,$ $\epsilon_j \sim U(-1,1)$ \\
- Normalize: $\pi_j = \frac{\pi_j}{\sum_{j=1}^{N}\pi_j}$\\
\item Choose question $q_s$ randomly, with respect to question weights $\pi_j$
\item return $q_s$
\end{enumerate} 
{\bf{Step 3: Update Question Grade:}}
\begin{enumerate}[topsep=0pt]
\setcounter{enumi}{4}
\item Get student grade $g_s$ after solving question $q_s$
\item reward $R = g_s-\eta$
\item {\bf{if}} $g_s>\eta$ {\bf{then}} 
\begin{itemize}
    \item Increase weights for questions more difficult than $q_s$: 
    \\ $w_j = \alpha_1 e^R w_j,$ for $j = s+1,\ldots,N$
	\item Increase exploration factor: $\gamma = \alpha_2 \gamma$ 
\end{itemize}
\item {\bf{else}}
\begin{itemize}
	\item Decrease weights for questions more difficult then $q_s$:\\
	$w_j = \alpha_3 e^R w_j,$ for $j = s+1,\ldots,N$
	\item Decrease exploration factor: $\gamma =  \alpha_4\gamma$
\end{itemize}
{\bf{end if}}
\item For each $j=1, \ldots , N$:\\
- Normalize: $w_j = \frac{w_j}{\sum_{j=1}^{N} w_j}$\\
\end{enumerate}
\caption{The Maple Approach}
\label{alg:MAPLE}
\end{algorithm}

Algorithm~\ref{alg:MAPLE} formalizes the MAPLE approach. 
The input to the algorithm is a set of students $S$ each with known solutions over a set of questions  in $Q$, a target student $s_i$ and a set of initialization parameters which include $\eta$, a passing grade threshold, $\gamma$, the exploration factor, and  $\alpha_1,\ldots,\alpha_4$, normalization factors. The algorithm returns the next question $q_s$ to present to student $s_i$.

The algorithm includes 3 steps: (1) The initialization step is performed once at the beginning of execution to obtain a personalized difficulty ranking for each student. During initialization the algorithm computes a difficulty ranking over questions per student. We used the EduRank approach for this purpose and describe it in the next section. Next, the question weights $w_j$ are initialized with values using a softmax function, with higher weights corresponding to easier questions per the difficulty ranking. As students solve questions and succeed or fail, these weights are updated for each student.

(2) Next Question Selection: performed at each time step by the agent to choose the next question to present to the student.
For this step MAPLE uses the distribution weights computed for the student. The algorithm first adds an exploration component per weight $w_j$ (line 2) generating new weights $\pi_j$, and then chooses the next question with respect to the  $\pi_j$ weights (lines 3,4).

(3) Update Question Grade: performed at each time step after the question was presented to the student and her solution grade was obtained. 
In line 6, a reward value $R$ is computed to reflect the magnitude of student's success or failure in the question with respect the the threshold value $\eta$. In line 7, the case of a successful solution is treated ($grade>\eta$): in this case, the exploration factor is increased, and weights of harder questions for this student are also increased proportionally to the reward value $R$. Thus, the probability of the student to get harder questions on next attempts increases, as well as the algorithm willingness to \mydoubleq{take risks} (exploration factor). 
On the final step, 9, weights are re-normalized on the unit scale.  
To summarize the update step, MAPLE weights update reflects the change in the algorithm's estimation as to the suitability of each question for the student based on expert guidelines. Nonetheless, the stochastic nature of the algorithm enables exploration of additional sequencing alternatives. We now move to describe our simulation and field trial results.

\section{Simulations with Synthesized Data}
We performed a set of simulations to compare four different sequencing algorithms:

(1) The \emph{MAPLE} approach which used the EduRank~\cite{segal2014edurank} algorithm for difficulty ranking. EduRank considers the history of students' actions in the system (including their grades and retries) and uses collaborative filtering~\cite{sarwar2001item} and voting aggregation approaches~\cite{nurmi1983voting} to compute a personalized difficulty ranking over questions. MAPLE's parameters for simulations were set empirically except for $\eta$ (the passing grade threshold) which was determined by an educational expert. 

(2) The \emph{Ascending} approach sequenced questions according to an absolute difficulty ranking that was determined by pedagogical experts. Questions were labeled into one of 5  groups (from easy to hard) according to  difficulty level. The algorithm selects questions in the following temporal order: The first 10\% of questions presented to students are level 1 questions (easiest level), followed by 20\% questions from level 2, 30\% questions from level 3, 30\% questions from level 4 and 10\% questions from level 5 (hardest level). This is the main sequencing approach used to sequence questions to students in the e-learning system tested in a school in the next section. 

(3) The  \emph{EduRank} approach provided a personalized difficulty ranking over questions for each student. Questions were sequenced from EduRank's easiest estimated question to its hardest estimated question per student.

(4) The \emph{Naive Maple} approach 
sequenced questions using the multi-armed bandit algorithm with random weights initialization  (without the EduRank based difficulty ranking component).

The algorithms were evaluated by comparing their performance in simulation. We model questions in the simulation using a $\langle$skill,difficulty$\rangle$ pair; students are modeled as a vector of skill values for each question type.

We now describe the details of the student model during simulation. 
The probability of a student to successfully solve a question is based on Item Response Theory~\cite{hambleton1991fundamentals} and is as follows:
 \begin{equation}
p(success) = \frac{1}{1+e^{\theta\cdot(ql - sl)}}
 \label{eq:1}
 \end{equation}
where $\theta$ is a constant influencing the shape of the probability distribution. The student's probability of learning is based on the difference between her skill level $sl$ and the level of the question $ql$.
To reflect the fact that students may differ in their answers due to factors beyond their skill level (e.g. guessing and slipping, affect condition) we add a stochastic component to Equation \ref{eq:1}. Thus, our student model behaves according to: 
 \begin{equation}
  p(success) = \beta\cdot\frac{1}{1+e^{\theta\cdot(ql - sl)}}+(1-\beta) \cdot \epsilon_u
 \label{eq:2}
 \end{equation}
 Where $\beta$ is a constant controlling the impact of the stochastic component and $\epsilon_u$ is drawn from a uniform distribution in the range [0,1]. 

Lastly we describe how the student skill level is updated upon solving a question. Intuitively, the skill level improves significantly following a correct response to a hard question when the student had a low level in that skill before answering the question. Similarly, the skill estimation was probably most exaggerated when the student failed in solving an easy question while having a high skill level before answering the question. 
    
This leads to the following skill update rules: If the student is successful in solving the question, i.e. her grade is greater than predefined value $\eta$, her skill level is updated according to: 
 \begin{equation}
  sl = sl+\delta_1\cdot ql \cdot(1-sl) 
 \label{eq:3}
 \end{equation}
 Otherwise, her skill level is updated according to:
  \begin{equation}
  sl = sl-\delta_2\cdot (1-ql)\cdot sl 
 \label{eq:4}
 \end{equation} 
The parameters $\theta, \beta, \delta_1, \delta_2$ are determined empirically. The parameter $\eta$ is determined by an educational expert and was set to 0.7. 

In the simulations each algorithm was run with a group of 1000 students, and was required to sequence 200 questions for each student. Each question belonged to one of 10 possible skills, uniformly distributed. We generated questions according to an absolute  difficulty level, uniformly distributed between  1 (easiest), and 5 (hardest). Students' initial competency level in each skill were also uniformly distributed between 0 (no skill knowledge) and 1 (full knowledge of skill). All the sequencing algorithms had access to \mydoubleq{historical} data generated by the simulation engine in a pre-simulation step so they can build their internal models. This data contained 1000 students each solving 500 randomly selected questions. 

\begin{figure}[H]
\includegraphics[width=1.2\textwidth,center]{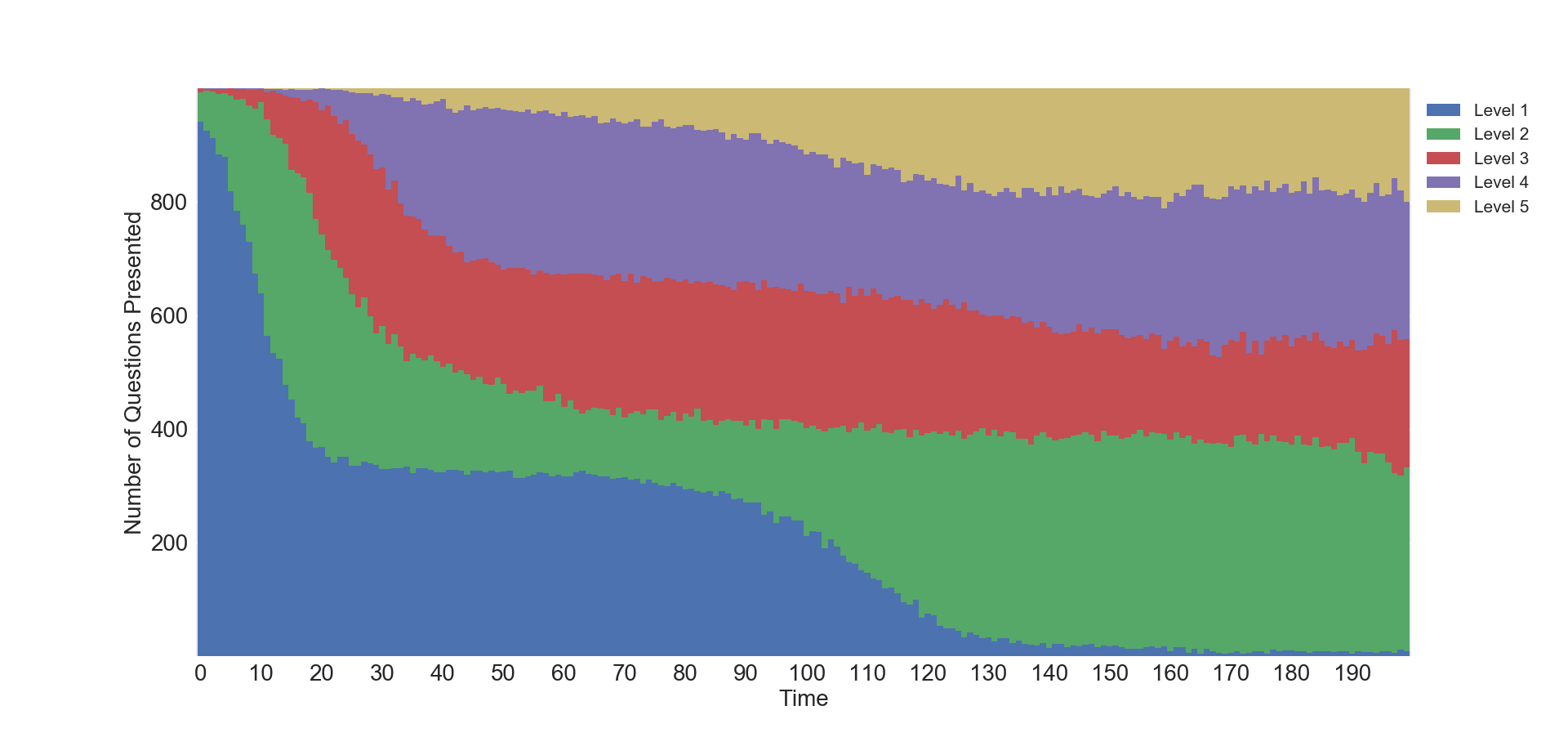}
\caption{MAPLE questions sequencing}
\label{fig:seqalgs}
\end{figure}

\begin{figure}[H]%
    \centering
    \subfigure[Weak Students]{{\includegraphics[width=10cm]{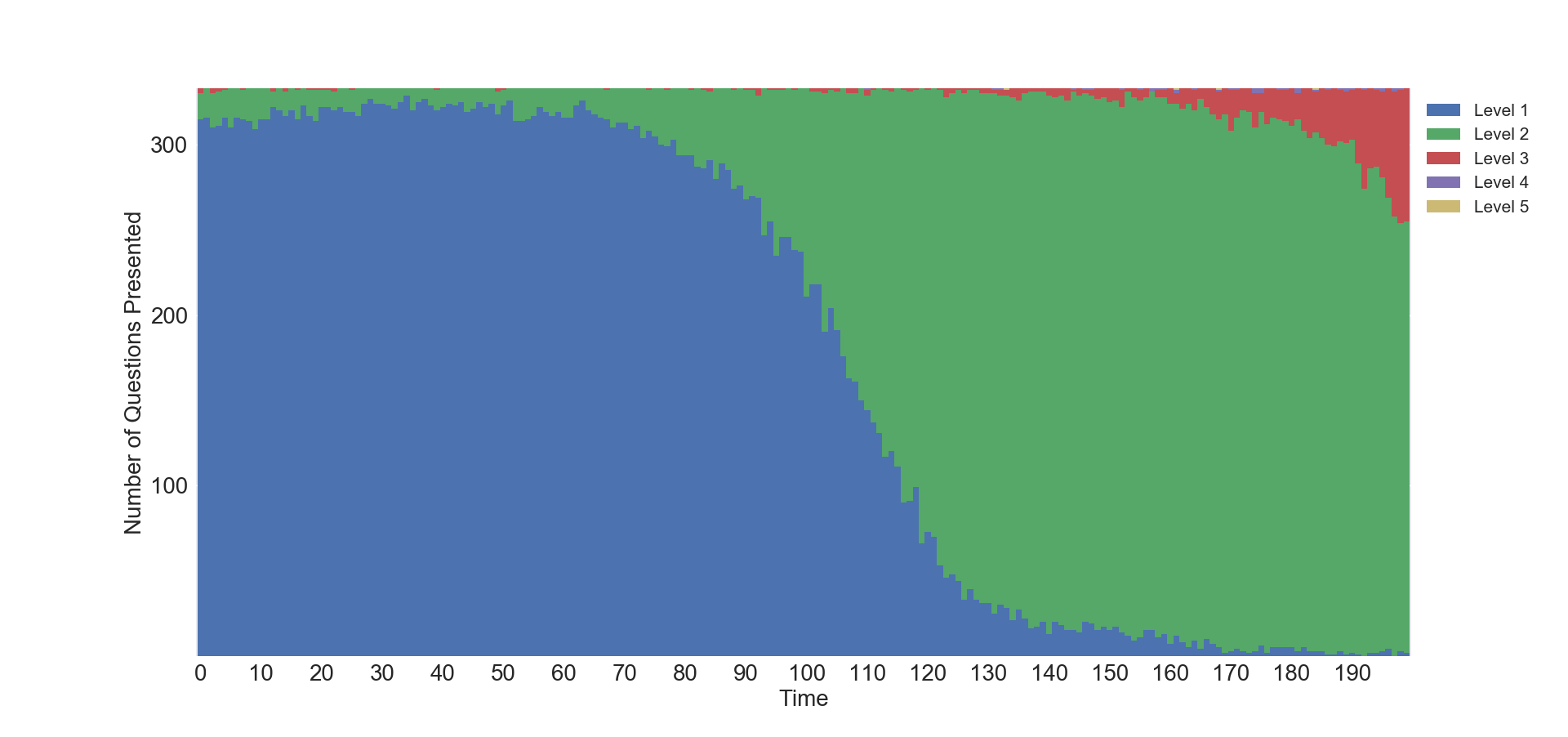} }}%
    \qquad
    \subfigure[Average Students]{{\includegraphics[width=10cm]{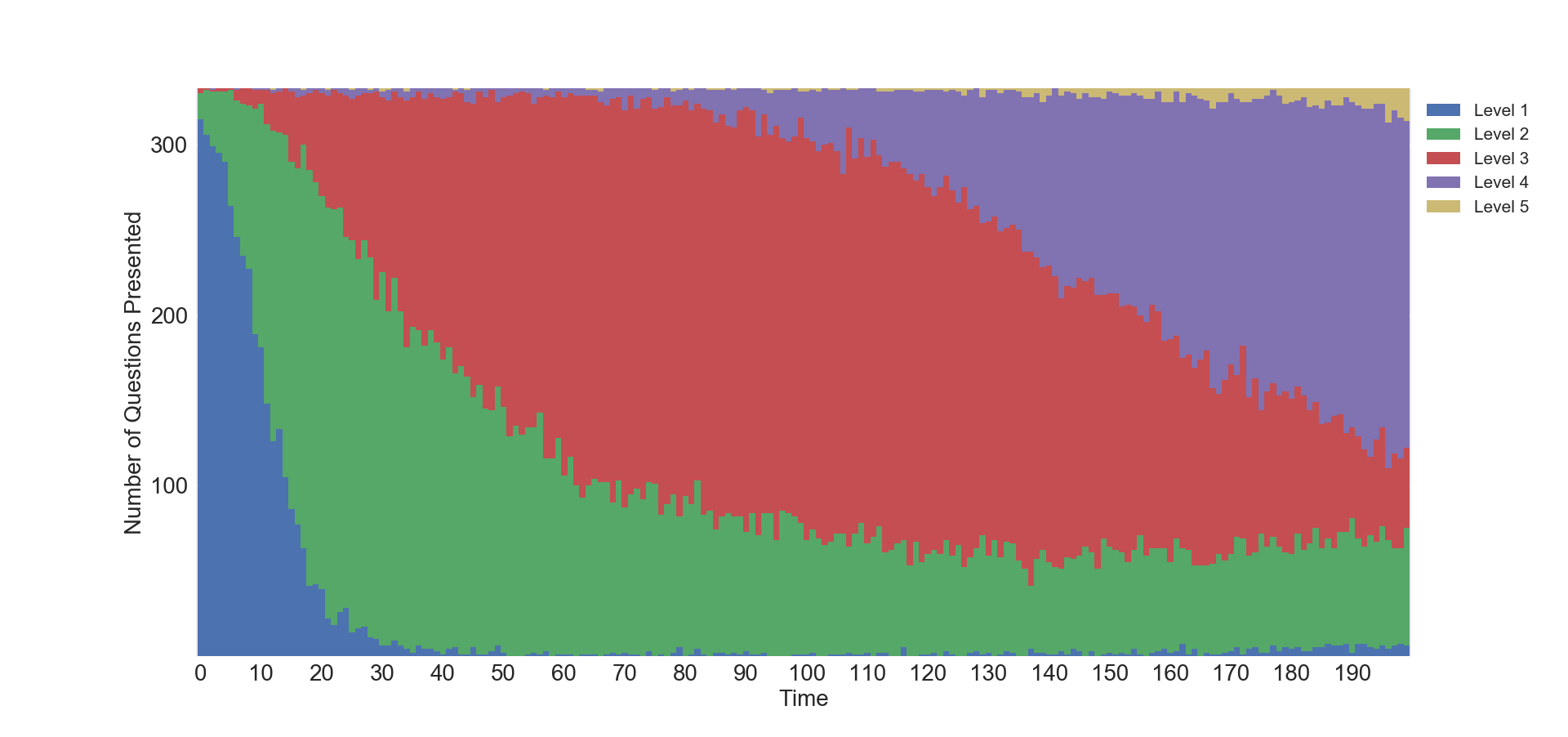} }}%
    \qquad
    \subfigure[Strong Students]{{\includegraphics[width=10cm]{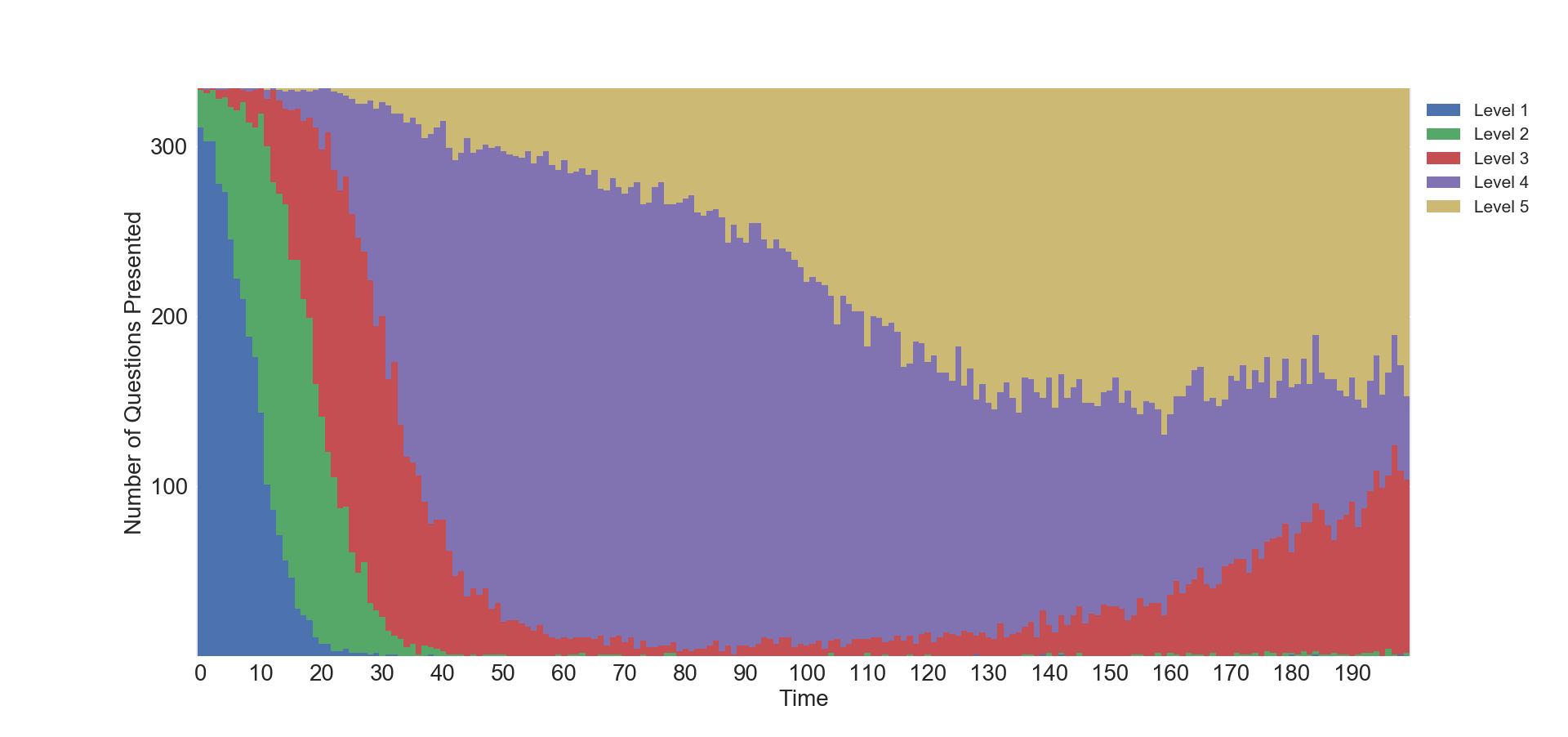} }}%
    \caption{MAPLE's question sequencing for different student types.}%
    \label{fig:simseq}%
\end{figure}

\begin{figure}
\centering
\includegraphics[width=1.4\textwidth,center]{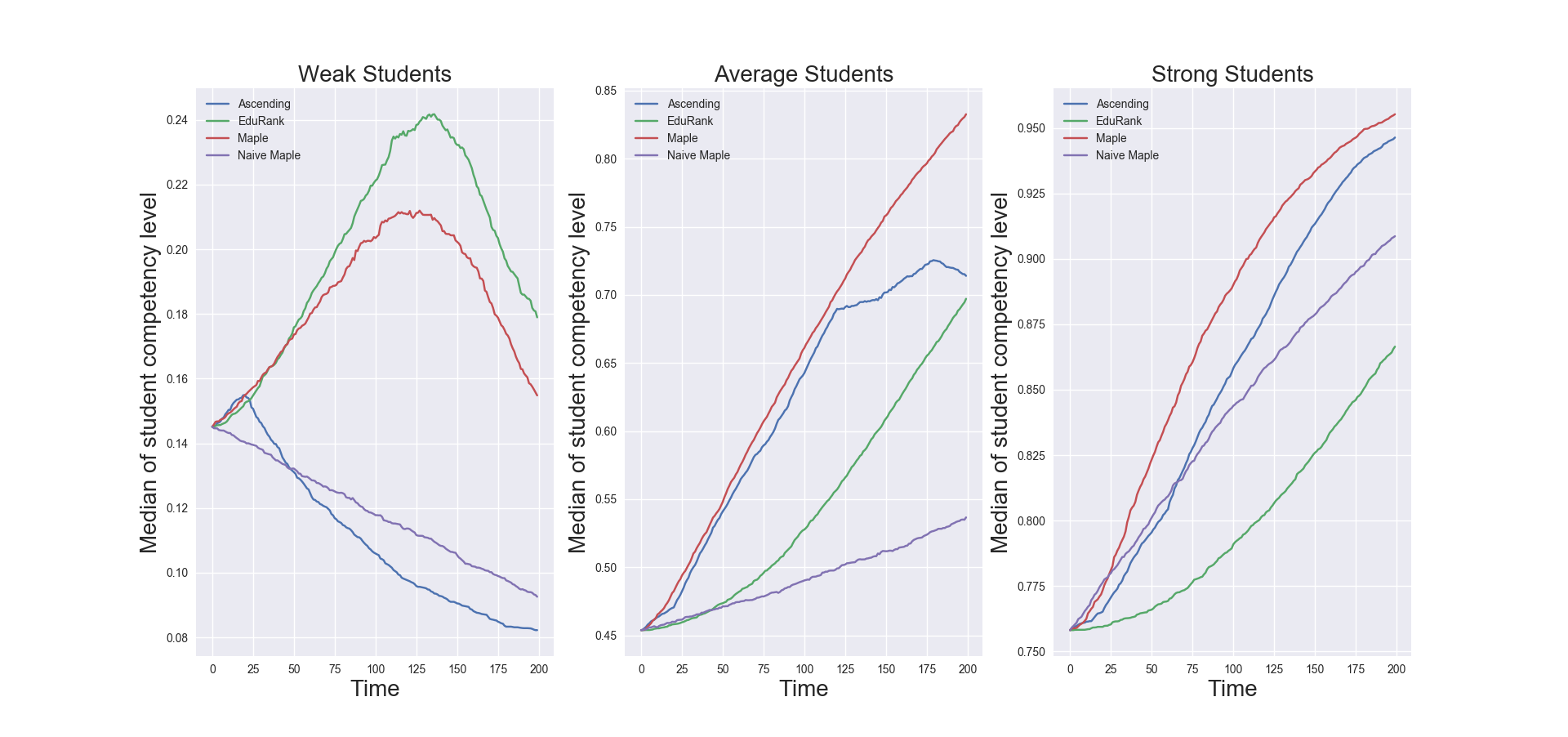}
\caption{Skill level progression per algorithm and student type.}
\label{fig:skillprog}
\end{figure}

We now describe key simulation results.
We start by looking at MAPLE's sequencing behaviour. Figure~\ref{fig:seqalgs} shows how MAPLE adapted the question difficulty as time progressed. The x-axis presents the time and the y-axis presents the number of questions of a given difficulty level that were presented to students in that time frame. The colors represent five difficulty levels ranging from easy (level 1) to hard (level 5). We can see that MAPLE started with offering easy questions to most students and then moved to propose more difficult questions while at the same time continuing to propose easy questions for a long time. This behaviour demonstrates the adaptive nature of MAPLE and was not observed in the other conditions tested. Specifically, EduRank used mostly easy questions and was not able to adapt harder questions to students while Naive Maple did not exhibit adaptive behavior for the questions in the practice set. (For longer practice sets the algorithm may show adaptive behavior).
To better understand MAPLE's adaptation behavior we next look at 3 different student groups: weak students (with initial skill competency levels under $0.33$), average students (with initial skill competency levels in the range [0.33, 0.67]), and strong students (with initial skill competency levels above $0.67$). 
Figure~\ref{fig:simseq} shows MAPLE's behavior for the 3 types of students and the 5 types of questions. We can see that for weak students, MAPLE kept proposing easy questions for a long time and refrained from proposing hard questions. For strong students, MAPLE begun proposing more difficulty questions earlier while reaching the hardest questions during the 200 questions session. And for average students, MAPLE took a middle way approach focusing on questions with average difficulty level while challenging some students with more difficult questions. 
Finally, we look at the skill level progression during simulation for the 3 student types in the 4 algorithms tested. 
Figure~\ref{fig:skillprog} shows the skill level progression of the different student types at each simulation step, for each algorithm.  We can see that for strong and average students MAPLE outperformed the other three algorithms throughout the practice session. Both EduRank which relies only on the difficulty information and Naive Maple which relies only on the multi-armed bandit approach presented lower results. For weak students we can see that the Ascending and Naive Maple approaches failed altogether, probably since they offered questions that are too hard for these students. Both MAPLE and EduRank presented initial good progress but then experienced a decline in the estimated skill level. This implies that MAPLE's adaptation scheme needs to be improved for this segment of students to offer less challenging questions. 

\section{Deployment and Evaluation in the Classroom}

We next moved to conduct a field study in the wild where students used  different sequencing approaches as part of their curriculum work in class (and not in a laboratory setting). 
The  MAPLE algorithm was implemented in an e-learning system used for Math education. In this system, K-12 students practice solving Math questions in various skill areas matching their curriculum studies. The study compared MAPLE in one school with 7th grade students, to two other existing sequencing algorithms already available in the e-learning system. The experiment was conducted between May 9th 2017 and June 19th 2017 (end of school year).
The students were randomly divided into 3 cohorts: (1) MAPLE Sequencing: Students in this group received questioned sequenced by the MAPLE algorithm when practicing with the system. (2) YBKT Sequencing\footnote{We note that the YBKT code was not available to us for the simulations.}: students in this group received questioned sequenced by the Bayesian Knowledge Trace based algorithm proposed  by Ben David et al. \cite{david2016sequencing}. This sequencing approach always chooses the next question from the available skill set in a deterministic manner. (3) Ascending Sequencing: students in this group received questions sequenced by the Ascending algorithm described earlier. 

All students in the experiment were initially exposed to a pretest session. In this session they solved 10 pretest questions that were hand picked by a pedagogical expert. The hand picked questions matched the expected level of 7th graders at that stage in the academic year. Ninety two students solved the pretest questions and there was no statistically significant difference between the three groups in the average score on this preliminary test. We thus concluded that the students in each group exhibited similar knowledge baselines of the material at hand. 

The students then engaged in multiple practice sessions in the e-learning system for the next 35 days, solving 10 assignment questions at each such practice session. For each cohort, assignment questions were sequenced by the cohort's respective algorithm (i.e. MAPLE, YBKT or Ascending). At the end of this period, students were asked to complete a post test session, solving the same questions (in the same order) as in the pretest session. Twenty eight students completed the post test session. We attribute the decrease in students' response from pretest to post test to the pending end of the academic year (there was no difference in the dropout rates across the 3 cohorts).

\begin{table}[H]
\centering
\scalebox{1.1}{
\begin{tabular}{|l|c|c|}
 \hline
Cohort & Time per Question (sec) & Average Grade \\
\hline
Ascending & 6.49 & 43.76 \\
MAPLE & 10.69 & 71.28 \\ 
YBKT & 12.86 & 67.08 \\
 \hline
 \end{tabular}
}
 \caption{Post test results per cohort: time per question and average grade.}
\label{tab:posttest}
\end{table}
Breaking up results by the sequencing algorithms, Table~\ref{tab:posttest} shows students' average grade and time spent on post-test questions, while Table ~\ref{tab:pre2post} shows the pre- to post- test change. As measured by post test grades, students assigned to the MAPLE condition achieved higher post test results than students assigned to the Ascending condition or to the YBKT condition.
This effect is also evident when observing the  pre- to post- test change in grades. Students assigned to the MAPLE condition learned  more than students assigned to the Ascending condition or to the YBKT condition. We note that further field trials with larger student groups are needed to evaluate statistical significance. 

In addition to objective measures of learning, we examined students' satisfaction from interacting with the various sequencing approaches. Students rated their degree of agreement on a 3 point scale ((1) I do not agree (2) I partially agree (3) I strongly agree) with ten statements about issues like the system helpfulness, ease of use, adaptivity, match to class material, and future appeal. 
Students using the MAPLE sequencing algorithm demonstrated higher satisfaction as represented by their answers to the subjective experience questions. The average score on the agreement with positive characteristics of the system was significantly different across the three conditions, with the highest score for the MAPLE algorithm at 2.39, compared to the Ascending algorithm (2.23) and the YBKT algorithm (2.20) (one way Anova, $p<0.05$). 
\begin{table}[H]
\centering
\scalebox{1.1}{
\begin{tabular}{|l|c|c|}
\hline
Cohort & Time per Question Diff & Average Grade Diff \\
\hline
Ascending & -9.2 & -2.83 \\
MAPLE & -9.12 & 4.99 \\
YBKT & -8.65 & 1.26 \\
\hline
\end{tabular}
}
\caption{Change from pre-test to post-test: time per question and average grade.}
\label{tab:pre2post}
\end{table}
\section{Conclusion}
We have provided a new computational method called MAPLE for sequencing questions to students in on-line educational systems. MAPLE combines difficulty ranking based on past student experiences with a multi-armed bandit approach using these difficulty rankings in on-line settings. 
We tested our approach in simulations and verified its adaptive nature and its learning gain results compared to other algorithms as measured by the simulated skills' improvement.  
We then performed a live experiment in the wild running MAPLE in a classroom in parallel to two baseline algorithms. We found that our proposed approach presented promising  results in students' performance and satisfaction.  
  
We mention several limitations of our work and subsequent suggestions for future work. First, the simulation results showed that MAPLE's performance needs to be fine tuned for weaker students. We plan to further investigate this issue in a followup research. Second, our student simulation  model is IRT based~\cite{hambleton1991fundamentals}  with assumptions taken about the behaviour of  skill progression. While these assumptions follow past work (see ~\cite{clement2013multi}), one can consider other skill progression assumptions as well as other models for student simulation (e.g. Performance Factor Analysis~\cite{pavlik2009performance}). We plan to extend the simulation code with other students' models. 
Finally, we had a small number of students performing the post test in the field experiment and plan to run larger scale field trials in future work to verify significance. 
 
\bibliographystyle{plain}
\bibliography{EMAB}
\end{document}